\documentclass[a4paper, 10pt, conference]{ieeeconf} 

\IEEEoverridecommandlockouts
\usepackage{tikz}
\usetikzlibrary{positioning, arrows.meta}
\usepackage{amsmath,amssymb,amsfonts}
\usepackage{algorithmic}
\usepackage{graphicx}
\usepackage{textcomp}
\usepackage{xcolor}
\usepackage{float}
\usepackage{orcidlink}
\usepackage[
style=ieee,
]{biblatex}

\def\BibTeX{{\rm B\kern-.05em{\sc i\kern-.025em b}\kern-.08em
    T\kern-.1667em\lower.7ex\hbox{E}\kern-.125emX}}

\DeclareMathOperator*{\argmax}{arg\,max\,}

\newcommand{\tho}{\boldsymbol{\theta}}
\newcommand{\thn}{\boldsymbol{\theta'}}
\newcommand{\policy}[1]{\pi_{{#1}}(\mathbf{a}|\mathbf{s})}
\newcommand{\expect}[2]{\mathbb{E}_{{#1}}\left[{#2}\right]}
\newcommand{\KL}{D_\text{KL}\left(\policy{\tho}||\policy{\thn}\right)}

\newif\ifanonym
\anonymfalse  % for final version with full repo link

\newcommand{\fullrepo}{https://github.com/jr-robotics/MorphologicalPPO}

\ifthenelse{\boolean{anonym}}{
  \newcommand{\repo}{\anonrepo}
}{
  \newcommand{\repo}{\fullrepo}
}

\begin{document}

\title{\LARGE \bf
Beyond Fixed Morphologies: Learning Graph Policies with Trust Region Compensation in Variable Action Spaces
}

\ifthenelse{\boolean{anonym}}{
    \author{Anonymous Author$^{1}$% <-this % stops a space
    \thanks{$^{1}$Anonymous Author is with Anonymous Institution, Anonymous Location
            {\tt\small anonymous.author@anonymousinstitution.com}}%
    }
}{
    \author{Thomas Gallien$^{1}$% <-this % stops a space
    \thanks{$^{1}$Thomas Gallien is with ROBOTICS – Institute for Robotics and Flexible Production, JOANNEUM RESEARCH Forschungsgesellschaft mbH, 9020 Klagenfurt, Austria
            {\tt\small thomas.gallien@joanneum.at\orcidlink{0000-0003-3331-5917}}}%
    }
}

\maketitle

%#################################################################################################################

\begin{abstract}

Trust region-based optimization methods have become foundational reinforcement learning algorithms that offer stability and strong empirical performance in continuous control tasks.
Growing interest in scalable and reusable control policies translate also in a demand for \textit{morphological generalization}, the ability of control policies to cope with different kinematic structures. Graph-based policy architectures provide a natural and effective mechanism to encode such structural differences. However, while these architectures accommodate variable morphologies, the behavior of trust region methods under varying action space dimensionality remains poorly understood. To this end, we conduct a theoretical analysis of trust region-based policy optimization methods, focusing on both Trust Region Policy Optimization (TRPO) and its widely used first-order approximation, Proximal Policy Optimization (PPO). The goal is to demonstrate how varying action space dimensionality influence the optimization landscape, particularly under the constraints imposed by KL-divergence or policy clipping penalties. Complementing the theoretical insights, an empirical evaluation under morphological variation is carried out using the Gymnasium Swimmer environment. This benchmark offers a systematically controlled setting for varying the kinematic structure without altering the underlying task, making it particularly well-suited to study morphological generalization.
\end{abstract}

\section{Introduction}

Deep reinforcement learning (DRL) has demonstrated remarkable success in solving continuous control tasks across various domains, including locomotion, manipulation, and autonomous navigation~\cite{Lillicrap2016aaa, Mnih2016aaa}. Among model-free algorithms, actor-critic methods such as Soft Actor-Critic (SAC)~\cite{Haarnoja2018aaa} or Trust Region Policy Optimization (TRPO)~\cite{Schulman2015aaa} have proven particularly effective. In particular, Proximal Policy Optimization (PPO)~\cite{Schulman2017aaa}, a simplified and more practical variant of TRPO, is widely adopted in the robotics community due to its ease of implementation and robust performance. PPO avoids second-order differentiation by introducing a clipped surrogate objective controlled by a clipping parameter $\epsilon$. In practice, this pragmatic approach of stabilization serves a purpose similar to the KL-constrained objective used for TRPO but requires much less computational effort. Studies on theoretical foundations have shown that, under certain assumptions, the clipped surrogate objective still attains global optimality, justifying practical success~\cite{Huang2024aaa}.

However, the question of how well policies generalize remains a central challenge in reinforcement learning. Meta-reinforcement learning has emerged as a prominent framework to address this issue by enabling agents to quickly adapt to novel tasks. Yet, the majority of meta-RL research focuses on generalization across tasks or reward structures, while implicitly assuming a fixed agent morphology. This assumption limits applicability in robotic domains, where robots often differ in kinematic structure. For example, a manipulation task like pick and place should ideally be solvable regardless of whether the agent controls a 6 or 7 DoF articulated robot equipped with a simple gripper or an anthropomorphic hand. In such cases the task remains essentially the same, although the control strategy must account for substantial differences in underlying kinematic architecture. This gives rise to the notion of morphological generalization, the ability to generalize across agents with varying kinematic structures, paving the way for scalable and transferable control policies.

One promising approach to enable morphological generalization is to leverage the structural inductive biases of Graph Neural Networks (GNNs). The structural composition of a robot takes the form of a kinematic tree, in which links and joints define nodes and edges in the graph representation, respectively~\cite{Wang2018aaa}. GNNs are best suited to operate on such relational data, as they can model interactions between components while remaining agnostic to the overall size or topology of the graph. This is particularly meaningful to robotic systems, where links naturally serve as storage for kinetic energy and joints define dynamic states through their relative configuration. From this perspective, message passing can be interpreted as modeling the flow of information and, implicitly, energy through the system, aligning closely with the physical realities of robotic control. This makes GNNs ideal candidates for representing policies that must generalize across morphologies. By embedding the agent's structure into the learning process, GNN-based policies can scale to a wide variety of robotic bodies, enabling zero-shot or few-shot generalization across agents with different numbers of links and joints and thus varying degrees of freedom. While empirical work has demonstrated the potential of GNNs in this context, theoretical understanding of how learning dynamics behave under such structural variability remains limited.

While GNN-based policies offer a natural mechanism for encoding kinematic structure, their successful application hinges on the learning algorithm’s ability to handle structurally diverse data. In practice, this entails training on batches of trajectories collected from morphologically different environments, each with potentially varying action space dimensionality. 
However, common actor-critic methods such as TRPO or PPO were originally designed under the assumption of a fixed action space. This raises a critical question: to what extent do these algorithms remain effective when applied to data aggregated across agents with heterogeneous morphologies? Surprisingly, the implications of such variability on policy optimization remains poorly understood, despite their increasing relevance in multi-robot and modular robotics settings.

This paper addresses this gap by analyzing how structural differences affect learning dynamics, with particular emphasis on the role of PPO's clipping mechanism when operating over variable action spaces. To this end, we conduct a theoretical analysis of trust region-based policy optimization methods, focusing on both TRPO and PPO. The goal is to determine how the dimensionality of the action space influences the updates to the network parameters, particularly under the constraints imposed by KL divergence and clipping-induced penalties. Complementing the theoretical insights, we perform an empirical evaluation using a modified version of the Gymnasium Swimmer environment. This setting is ideally suited for studying morphological generalization, as its kinematic structure can be systematically varied without altering the core of the locomotion task. By leveraging this controllability, we investigate how policy learning is affected by morphological diversity and assess the robustness of PPO in such settings. Code supporting this work is available at \url{\repo}.

\section{Related Work}
\label{sec:related_work}

\subsection{Adaptive Clipping}
The clipping mechanism plays an important role in the algorithm's learning performance of and has therefore been the subject of numerous studies focusing on enhancing PPO's robustness and adaptability. Evaluation studies on hyperparameter tuning have revealed that $\epsilon$ is one of the most sensitive parameters and must be carefully considered due to its significant impact on training stability and overall performance~\cite{Engstrom2020aaa}. Vanilla PPO implementations use constant clipping thresholds, which is disadvantageous in some situations. In contrast, adaptive clipping methods aim to modify clipping behavior based on dynamic strategies by adjusting thresholds based on observed distributions of policy updates~\cite{Chen2018aaa} or gradually decaying clipping ranges as convergence approaches~\cite{Farsang2021aaa}. Han et al.~\cite{Han2019aaa} introduced dimension-wise clipping to reduce bias in high-dimensional action spaces by applying per-dimension constraints. Other methods integrate statistical modeling, such as heavy-tailed gradients~\cite{Garg2021aaa} or upper-confidence bounds via bandit-style control~\cite{Zhang2023aaa, Massaoudi2025aaa}, improving robustness in uncertain environments.

\subsection{Alternative Mismatch Penalties}
While most trust region-based methods such as TRPO and PPO constrain policy updates via KL-divergence penalties or clipping, alternative discrepancy measures have been proposed to improve stability, exploration, and generalization. One prominent direction replaces the KL constraint with distances derived from optimal transport theory, particularly the Wasserstein metric. Policy optimization as a Wasserstein gradient flow was introduced in~\cite{Zhang2018aaa}, providing a principled alternative to KL-based trust regions. Building on this idea, a TRPO variant with optimal transport discrepancies was developed in~\cite{Terpin2022aaa}, including dual formulations and efficient algorithms for continuous actions. Wasserstein gradient flows have also been applied to Gaussian mixture policies~\cite{Ziesche2023aaa}, enabling more expressive and multimodal action distributions, and functional Wasserstein variational policy optimization was proposed in~\cite{Xuan2024aaa} to capture richer functional constraints in policy updates. The framework was further generalized in~\cite{Pfau2025aaa}, demonstrating benefits in large-scale deep RL settings. Orthogonal to these algorithmic contributions, differentiable trust region layers were introduced in~\cite{Otto2021aaa}, allowing integration of alternative penalties directly into neural architectures for end-to-end learning.

\subsection{Action Space Modification}
Structured action spaces pose challenges for trust region-based policy optimization, including inefficient exploration, unstable updates, and poor convergence. To mitigate these issues, methods have emerged to reshape or constrain action spaces for improved learning. Discretizing continuous spaces, for instance, lowers gradient variance and introduces structured exploration, helping avoid unstable updates in high-dimensional domains. Ordinal parameterization further incorporates inductive bias by encoding natural ordering in discrete actions~\cite{Tang2020aaa}. Action space shaping simplifies decision-making by restructuring available actions~\cite{Kanervisto2020aaa}, while action masking prevents the policy from selecting infeasible or suboptimal actions. In discrete domains, masking improves sample efficiency by eliminating invalid choices~\cite{Huang2O22aaa}. Recent work extends masking to continuous settings, refining learning by restricting policies to task-relevant regions—especially effective in structured locomotion tasks and morphologically diverse agents~\cite{Stolz2024aaa}.

\subsection{Graph-Structured Policies}

Early foundational work introduced GNNs for learning physics simulation engines, demonstrating their capability to capture relational information essential for prediction and control~\cite{Sanchez-Gonzalez2018aaa, Sanchez-Gonzalez2020aaa, Pfaff2021aaa}. Architectures such as NerveNet build upon that idea introducing GNN-based actor-critic agents learning morphologically diverse agents for locomotion control~\cite{Wang2018aaa}. Further advancements extended this approach to modular robotic systems, where GNNs-based policies were utilized to accommodate variable topologies dynamically~\cite{Khan2020aaa}. More recent developments explicitly model morphological heterogeneity, supporting generalization across agents with diverse kinematic and structural properties~\cite{Hao2024aaa}, where bi-level optimization utilizing GNN-policies have shown to offer zero-shot transfer capabilities~\cite{Gammelli2023}.

\subsection{Meta-Reinforcment Learning}
The broader goal of generalization is a is also central to the field of meta-reinforcement learning. Meta-RL traditionally emphasizes adaptation across different task distributions~\cite{Kirsch2020aaa}, yet a closely related challenge arises when solving structurally similar tasks across agents with varying kinematic configurations. Several recent works have begun to address this intersection by applying meta-learning techniques to the control of legged robots~\cite{Belmonte-Baeza2022aaa}, \cite{Song2020aaa}, where fast adaptation to novel morphologies or terrains is critical. However, most meta-RL approaches still treat morphology as fixed and focus on variability in external tasks, whereas the present work targets structural variation as the primary source of generalization, particularly with regard to optimizing trust region-based agents for mixed variable-dimensional data batches.

\section{Theoretical Analysis}
The present work builds upon previous studies incorporating GNN policies using Proximal Policy Optimization (PPO) to learn locomotion tasks with varying degrees of freedom~\cite{Wang2018aaa}. However, a fundamental question remains unanswered: To which extend does the varying dimension of the action space influence the learning algorithm? To answer this question we dedicate this section to lay ground for a theoretical analysis for both algorithms, TRPO and it's popular derivative PPO.

\subsubsection{Trust Region Policy Optimization}
We start the analysis with TRPO's objective as it serves as a basis for it's more easier to implement derivative PPO:
\begin{align}
    \hat{\thn} = &\argmax_{\thn} 
        \expect{\mathbf{s}, \mathbf{a}\sim\pi_{\tho}}{\frac{\policy{\thn}}{\policy{\tho}} A^{\pi_{\tho}}(\mathbf{s}, \mathbf{a})} \nonumber \\
    &\text{subject to} \nonumber \\
    &\expect{\mathbf{s}}{\KL} \leq \delta .\,\label{eq:trpo_objective} 
\end{align}
The usual approach to solving this problem is to approximate the objective function and the constraint using first- and second-order Taylor expansions at $\tho$, respectively, and then saturate the constraint to derive tractable quadratic optimization problem giving the Lagrangian:
\begin{align}
    \mathcal{L}(\Delta\boldsymbol{\theta}, \lambda) = 
        \mathbf{g}^T\Delta\boldsymbol{\theta}  - 
        \lambda\left(\frac{1}{2}\Delta\boldsymbol{\theta}^T\mathbf{F}_{\pi_{\tho}}\Delta\boldsymbol{\theta} - \delta\right) ,\, \label{eq:trpo_aug_lagrange}
\end{align}
where $\Delta\boldsymbol{\theta}$ denotes the update of the parameter vector, $\mathbf{g}$ is the advantage-weighted policy gradient and $\mathbf{F}_{\pi_{\tho}}$ is the Fisher information matrix of policy, respectively. Solving the Lagrangian~\eqref{eq:trpo_aug_lagrange} reveals the expression for the parameter update:
\begin{align}
    \Delta\boldsymbol{\theta}_\text{TRPO} = 
        \sqrt{\frac{2\delta}{\mathbf{g}^T\mathbf{F}_{\pi_{\tho}}^{-1}\mathbf{g}}}
        \mathbf{F}_{\pi_{\tho}}^{-1}\mathbf{g} ,\,\label{eq:trpo:delta_theta}
\end{align}
where the majority of implementations avoid calculating the inverse of the Fisher information and solve the problem by means of conjugate gradient optimization instead. Nevertheless, \eqref{eq:trpo:delta_theta} highlights the influence of the Fisher information on the parameter update and hence serves as the basis for further analysis. Since $\policy{\tho}$ is generally a factorized distribution, $\mathbf{F}_{\pi_{\tho}}$ decomposes into a sum over the Fisher information matrices of the individual marginal distributions. By definition, $\mathbf{F}_{\pi_{\tho}}$ must be positive definite to ensure invertibility and enable a well-posed solution for the parameter update vector. Consequently, we can assume that the eigenvalues $\lambda_i$ of $\mathbf{F}_{\pi_{\tho}}$ scale proportionally with the dimensionality of the action space, i.e., $\lambda_i\sim \mathcal{O}\left(\dim(\mathcal{A})\right)$.
Decomposition of the update vector in the eigenspace of $\mathbf{F}_{\pi_{\tho}}$ reveals the $L_2$-norm of the parameter update to be:
\begin{align}
    \|\Delta\boldsymbol{\theta}_\text{TRPO}\| = \sqrt{2\delta\frac{\boldsymbol{\alpha}^T\boldsymbol{\Lambda}^{-2}\boldsymbol{\alpha}}{\boldsymbol{\alpha}^T\boldsymbol{\Lambda}^{-1}\boldsymbol{\alpha}}} ,\,
\end{align}
where $\mathbf{F}_{\pi_{\tho}} = \mathbf{Q}\boldsymbol{\Lambda}\mathbf{Q}^T$ and $\boldsymbol{\alpha} = \mathbf{Q}^T\mathbf{g}$, respectively.
The only relevant contributor is the diagonal matrix of the eigenvalues since $\boldsymbol{\alpha}$ is not affected by the dimension of the action space. Hence, the ratio scales indirect proportional with the dimension of the action space and thus the $L_2$-norm of the parameter update according to:
\begin{align}
    \|\Delta\boldsymbol{\theta}_\text{TRPO}\| \sim \mathcal{O}\left(\frac{1}{\sqrt{\dim(\mathcal{A})}}\right) .\, \label{eq:L2_trpo}
\end{align}
Equation~\eqref{eq:L2_trpo} reveals that the magnitude of the TRPO parameter update contracts inversely with the square root of the action space dimensionality. Consequently, in mixed-batch scenarios processing data originating from multiple action space dimensions the optimization is biased towards lower-dimensional samples, which allow larger parameter updates and effectively over-regularizing higher-dimensional ones.

\subsubsection{Proximal Policy Optimization}
PPO builds upon the idea of trust region optimization and offers a more practical and scalable alternative. Schulman et al.~\cite{Schulman2017aaa} essentially introduced two versions, one closely related to the TRPO objective~(\ref{eq:trpo_objective}) and one avoiding second order differentiation by introducing an unconstrained nonlinear objective. These are commonly referred to as the \textit{$\beta$-surrogate objective} and the \textit{clipped-surrogate objective}, respectively.

We will start the analysis with the \textit{$\beta$-surrogate objective}, which essentially slightly modifies (\ref{eq:trpo_objective}) by replacing the constraint with a penalty controlled via the hyperparameter $\beta$. Similar to before we use first and second order Taylor expansion at $\tho$ to derive a unconstrained quadratic optimization problem:
\begin{align}
    \mathcal{L}(\Delta\boldsymbol{\theta}) = 
        \mathbf{g}^T\Delta\boldsymbol{\theta}  - 
        \frac{\beta}{2}\Delta\boldsymbol{\theta}^T\mathbf{F}_{\pi_{\tho}}\Delta\boldsymbol{\theta} ,\, 
\end{align}
and parameter update vector:
\begin{align}
     \Delta\boldsymbol{\theta}_{\text{PPO},\beta} = \frac{1}{\beta}\mathbf{F}_{\pi_{\tho}}^{-1}\mathbf{g} ,\,
\end{align}
accordingly. Relaxing the optimization problem evidently comes at a cost, as the parameter update is not scaled anymore according to the definition of a trust region due to the absence of the Lagrange multiplier. Unsurprisingly, this impacts also the influence of the of the action space dimension on the $L_2$-norm of the parameter update:
\begin{align}
    \|\Delta\boldsymbol{\theta}_{\text{PPO,$\beta$}}\| = 
        \frac{1}{\beta}\sqrt{\boldsymbol{\alpha}\boldsymbol{\Lambda}^{-2}\boldsymbol{\alpha}} \quad \sim\mathcal{O}\left(\frac{1}{\dim(\mathcal{A})}\right) ,\,
\end{align}
where the update vector is once again decomposed in the eigenspace of the Fisher information matrix. Apparently, the dimensional bias intensifies significantly, suggesting worse generalization capabilities of the $\beta$-surrogate objective with respect to varying action space dimensions. 

However, most PPO implementations avoid second order differentiation and aim to maximize the \textit{clipped surrogate objective} instead:
\begin{align}
    \mathcal{L}(\thn) = \expect{t}{\min\left(r_tA^{\pi_{\tho}}, \text{clip}\left(r_t, 1-\epsilon, 1+\epsilon\right)A^{\pi_{\tho}}\right)} ,\,
\end{align}
where $r_t(\thn) = \frac{\policy{\thn}}{\policy{\tho}}$ denotes the importance ratio of the updated compared to the sampled policy and $\epsilon\in \mathbb{R}^{+}$ is a small-valued hyperparameter and controls the tolerable mismatch similar to $\delta$ in case of TRPO. 
In contrast to the evaluation above, the influence of the action space dimension isn't directly visible via the Fisher information anymore.
Instead we aim to assess how the probability that $r_t$ remains unclipped is influenced by the dimension of the action space. 

The policy is in general a factorized distribution which also applies to the importance ratio $r_t = \prod_{i=1}^{\dim(\mathcal{A})} r_{i,t}$, where $r_{i,t}$ defines usually the ratio of two Gaussian marginals in case of a continuous control problem. Hence, the logarithm of the
i\textsuperscript{th}-marginal ratio is given by:
\begin{align}
    \log(r_{i,t}) = 
        \frac{1-\eta_i^{-2}}{2}z_i^2 +
        \frac{\nu_i}{\eta_i} z_i -
        \log(\eta_i) - \frac{1}{2}\nu_i^2 ,\,
\end{align}
where $\nu_i = (\mu_{i,\thn} - \mu_{i,\tho})/\sigma_{i,\thn}$ defines a normalized drift value and $\eta_i = \sigma_{i,\thn}/\sigma_{i,\tho}$ the ratio of standard deviation. $z_i$ corresponds to a standard normal distributed random variable since $\sigma_{i\tho} z_i + \mu_{i,\tho} = a_i \sim\pi_{i,\tho}$. Apparently, $\log(r_{i,t})$ is a quadratic function of a normal distributed random variable of the form $\alpha_i z_i^2 + \beta_i z_i + \gamma_i$, hence $\log(r_t)$ resembles approximately a non-centered $\chi$-squared distribution:
\begin{align}
    \log(r_t) = \sum_{i=1}^{\dim(\mathcal{A})}
        \frac{1-\eta_i^{-2}}{2}z_i^2 +
        \frac{\nu_i}{\eta_i} z_i -
        \log(\eta_i) - \frac{\nu_i^2}{2} .\,
\end{align}
However, in most PPO implementations, the log variance is modeled as a free parameter independent of the observations, and thus evolves only slowly across gradient updates. Consequently, we can assume $\eta_i\approx 1$ and thus $\log(r_t)$ approximately distributed according to a Gaussian distribution with mean $\mu_{\log{r_t}} = -\frac{1}{2}\|\boldsymbol{\nu}\|^2$ and standard deviation $\sigma_{\log{r_t}} = \|\boldsymbol{\nu}\|$.
We can now evaluate the probability that the (log)policy ratio falls into the interval $[1-\epsilon, 1+\epsilon]$ which corresponds approximately to $\mathbb{P}\left[-\epsilon \leq \log(r_t) \leq \epsilon\right]$ as $\log(1\pm\epsilon)\approx \pm\epsilon$:
\begin{align}
    \mathbb{P}\left[-\epsilon \leq \log(r_t) \leq \epsilon\right] = \nonumber \\
    \Phi\left(\frac{\epsilon}{\|\boldsymbol{\nu}\|} + \frac{1}{2}\|\boldsymbol{\nu}\|\right) - 
    \Phi\left(\frac{-\epsilon}{\|\boldsymbol{\nu}\|} + \frac{1}{2}\|\boldsymbol{\nu}\|\right) .\, \label{eq:true_prob}
\end{align}
First order local approximation of the Gaussian cdf simplifies the expression to:
\begin{multline}
    \mathbb{P}\left[-\epsilon \leq \log(r_t) \leq \epsilon\right] \approx
    \frac{1}{\sqrt{2 \pi}} \exp\left(-\frac{1}{8}\|\boldsymbol{\nu}\|^2\right)\frac{2\epsilon}{\|\boldsymbol{\nu}\|} .\,
\end{multline}
We argue that the exponential decay in the expression can be neglected under the assumption that $\|\boldsymbol{\nu}\|^2 \ll 1$ since $\boldsymbol{\nu}$ is primarily governed by the drift of the policy's mean parameters. This drift remains small in practice due to the use of trust region-inspired update constraints. Consequently, the dominant factor influencing the probability of staying within the clipping range is the inverse proportionality to $\|\boldsymbol{\nu}\|$, which scales as $\mathcal{O}(\sqrt{\dim(\mathcal{A})})$ with increasing action space dimensionality. Therefore, the probability that the policy ratio falls into the $\pm\epsilon$ interval can be approximated as:
\begin{align}
    \mathbb{P}\left[-\epsilon \leq \log(r_t) \leq \epsilon\right] \approx \min\left(\sqrt{\frac{2}{\pi}}\frac{\epsilon}{\|\boldsymbol{\nu}\|}, 1\right)
    \propto \frac{1}{\sqrt{\dim(\mathcal{A})}} .\, \label{eq:PPO_prob_logr}
\end{align}

\begin{figure}
    \centering
    \includegraphics[width=0.95\linewidth]{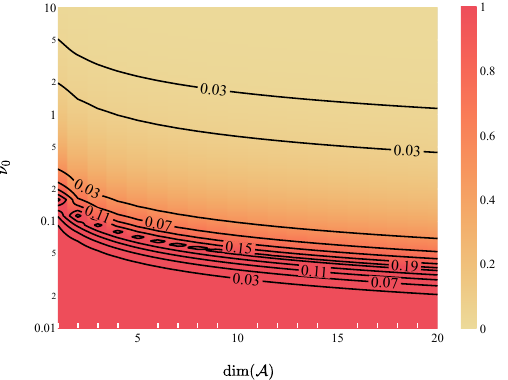}
    \caption{
    Probability $\mathbb{P}\left[-\epsilon \leq \log(r_t) \leq \epsilon\right]$ as a function of the action space dimensionality and the homogeneous normalized drift $\nu_0$, with clipping threshold $\epsilon = 0.2$. Contour lines indicate the approximation error with respect to the analytic expression $\min\left(\sqrt{\frac{2}{\pi}} \frac{\epsilon}{|\boldsymbol{\nu}|}, 1\right)$, compared against the true probability as derived in~(\ref{eq:true_prob}).
    }
    \label{fig:no_clip_prob}
\end{figure}

\begin{figure}
    \centering
    \includegraphics[width=0.95\linewidth]{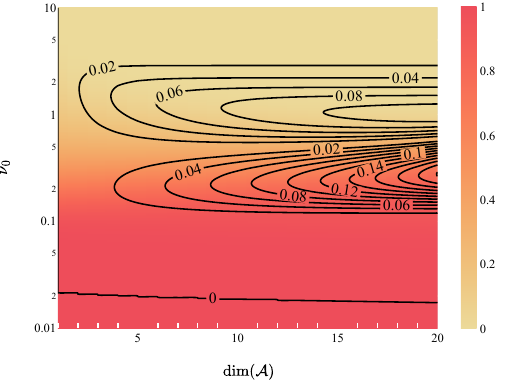}
    \caption{Compensated probability as a function of action space dimension and normalized drift $\nu_0$ for $\epsilon = 0.2$, where the effective clipping value is scaled by $\sqrt{\dim\left(\mathcal{A}\right)}$. Contour lines denote the approximation error relative to the reference case with $\dim(\mathcal{A}) = 1$.}
    \label{fig:no_clip_prob_comp}
\end{figure}

Fig.~\ref{fig:no_clip_prob} supports this assumption, where we analyzed the accuracy of the approximation in dependence of the action space dimension and normalized drift conditions for $\epsilon=0.2$. Contour lines illustrate the approximation error which reaches it's maximum value of 0.19 for higher dimensions when the evaluation takes place near the $\pm\sigma_{\log{r_t}}$ interval but isn't uniformly high elsewhere, indicating only localized regions of noticeable deviation.
This is more visible in Fig.~\ref{fig:no_clip_prob_comp} where we compensated the influence of the action space dimension by evaluating the distribution at $\pm\epsilon\sqrt{\dim(\mathcal{A})}$.

\section{Evaluation}

\subsection{Setup}
\begin{figure}[h]
    \centering
    \includegraphics[width=0.95\linewidth]{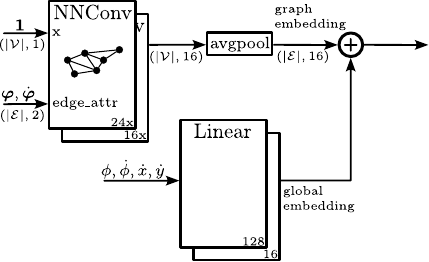}
    \caption{Schematic illustration of the GNN-based feature extractor. Observations associated with robot joints, such as joint angles and joint angular velocities ($\boldsymbol{\varphi}, \dot{\boldsymbol{\varphi}}$), are structured as a graph and processed using a two-layer edge-conditioned convolutional (NNConv) network with 24 and 16 channels, respectively. Additional global observations, including yaw ($\phi$), yaw velocity ($\dot{\phi}$), and linear velocities of the root body ($\dot{x}, \dot{y}$), are processed by a fully connected network (128 and 16 neurons) and subsequently fused with the graph feature embeddings. All hidden layers are ReLU activated.}
    \label{fig:arch}
\end{figure}

For our PPO implementation, we adapted code from Stable Baselines 3~\cite{stable-baselines3} to support variable-sized action spaces and used PyTorch Geometric~\cite{torch-geometric} as the framework for the policy. 
Actor and critic share the same feature extractor shown in Fig.~\ref{fig:arch} which is based on a two-layer edge-conditioned convolutional (NNConv) network that processes observations structured as a graph of robot joint states. Global observations, such as yaw and the linear velocity of the root body, are processed by a small fully connected network and fused with the graph embeddings. The feature extractor is highly lightweight, comprising only 4376 trainable parameters. This highlights the  expressive power of GNNs for controlling kinematic structures. 
As environment we used a modified version of Gymnasium's Swimmer~\cite{gymnasium} environment to enable adaptability to a variable number of joints. During training, the number of joints was randomly sampled according to $\dim(\mathcal{A})\sim \mathcal{U}\left[2,10\right]$. All other parameters were kept at their default values.

\subsection{Results and Discussion}
\begin{figure}[h]
    \centering
    \includegraphics[width=0.95\linewidth]{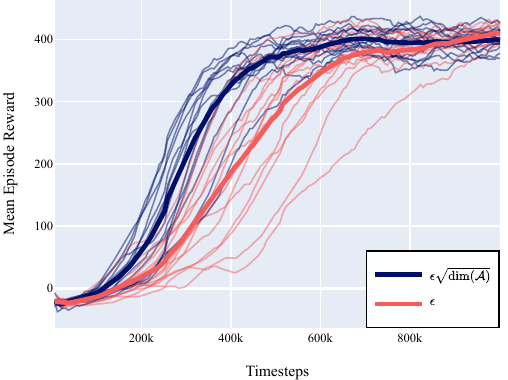}
    \caption{Training curves showing the evolution of mean episode rewards under two clipping schemes across different random seeds for $\epsilon=0.111$: the dimension-compensated limit ($\epsilon\sqrt{\dim(\mathcal{A})}$) and the standard uncompensated limit ($\epsilon$). Thick lines represent ensemble mean reward traces.}
    \label{fig:mean_reward_train}
\end{figure}

\begin{figure}[h]
    \centering
    \includegraphics[width=0.95\linewidth]{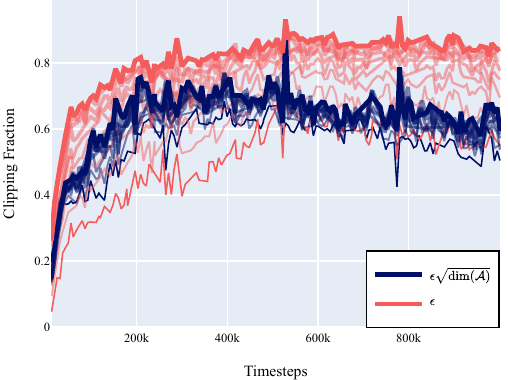}
    \caption{
        Visualization of clipping fractions per action space dimension during training, comparing dimension-compensated clipping ($\epsilon\sqrt{\dim(\mathcal{A})}$) with the uncompensated baseline ($\epsilon$). The line widths represent the action space dimension, starting from dimension 2 (thinnest) to dimension 10 (thickest). Semi-transparent lines indicate intermediate values. 
    }
    \label{fig:clip_train}
\end{figure}

To ensure a fair comparison between the dimension-compensated method and the uncompensated baseline, we optimized hyperparameters only for the former, focusing on the clipping parameter $\epsilon$ while keeping all other parameters at their default values. Fig.~\ref{fig:mean_reward_train} shows the mean episode reward during training for a clipping value of $\epsilon=0.111$ across different random seeds.
The results reveal that both configurations are capable of solving the task, achieving comparable final episodic rewards. However, the dimension-compensated configuration exhibits a markedly faster convergence, reaching the plateau reward level approximately 200k timesteps earlier than the uncompensated baseline. This suggests that adjusting the clipping threshold to account for action space dimensionality can improve sample efficiency by ensuring uniformly clipped gradients across different action dimensions. The clipping likelihood follows the expected trend as shown in Fig.~\ref{fig:clip_train}, with per-dimension fractions showing markedly lower spread in the compensated setting—consistent with theory and primarily due to approximation error when the clipping interval approaches $\pm\sigma_{\log{r_t}}$.

\begin{figure}[h]
    \centering
    \includegraphics[width=1\linewidth]{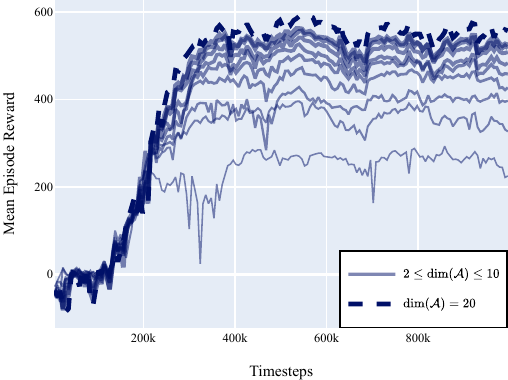}
    \caption{
        Evaluation results per action space dimension for policy snapshots taken every 6,000 timesteps during training for the dimension-compensated configuration. Line widths encode the action space dimension in ascending order. Semi-transparent solid lines correspond to dimensions present during training ($2\leq\dim(\mathcal{A})\leq10$). The dashed line shows generalization performance for $\dim(\mathcal{A})=20$.
    }
    \label{fig:mean_reward_val_varppo}
\end{figure}

\begin{figure}[h]
    \centering
    \includegraphics[width=1\linewidth]{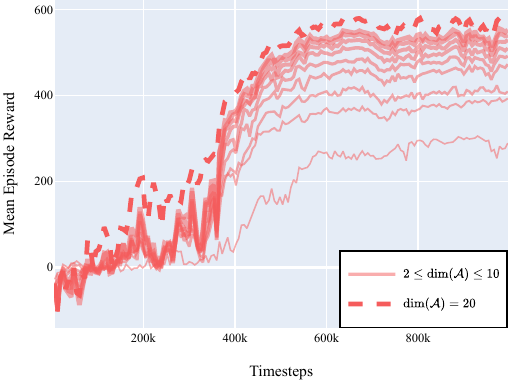}
    \caption{
        Evaluation results per action space dimension for policy snapshots taken every 6,000 timesteps during training for the uncompensated baseline. Line widths encode the action space dimension in ascending order. Semi-transparent solid lines correspond to dimensions present during training ($2\leq\dim(\mathcal{A})\leq10$). The dashed line shows generalization performance for $\dim(\mathcal{A})=20$.
    }
    \label{fig:mean_reward_val_ppo}
\end{figure}

For evaluation, we loaded policy snapshots every 6,000 timesteps and tested a 20-DoF environment in addition to the training configuration to assess generalization. As in training, the dimension-compensated configuration converged considerably faster than the uncompensated baseline. Surprisingly, both configurations achieved higher performance in higher-dimensional action spaces, with the untrained 20-DoF policy outperforming all others which is shown in Fig.\ref{fig:mean_reward_val_varppo} and Fig.\ref{fig:mean_reward_val_ppo}, respectively. This is counterintuitive, as clipping is generally more likely in higher dimensions, which should, in principle, reduce the amount of policy gradient information propagated to the parameter update. One would therefore expect worse performance in higher dimensions, with this effect being less pronounced for the compensated setting. However, analyzing the motion pattern of the environment provides an intuitive explanation: as the number of DoF increases, the Swimmer’s motion increasingly resembles a harmonic waveform. This reduces the required dynamics for each individual joint, as each joint covers a smaller range of movement per step compared to low-dimensional scenarios. Consequently, the policy’s marginal distributions can be modeled with less complexity, making them easier to train effectively.

\section{Conclusion \& Outlook}

This work presents, to the best of our knowledge, the first analytical investigation of how action space dimensionality affects trust region–based reinforcement learning algorithms. By deriving and empirically validating theoretical predictions, we provide a principled understanding of clipping behavior in PPO under varying action space sizes. In addition, we extend the analysis to the morphological dependence of graph neural network–based policies, examining how policy performance and convergence dynamics are influenced by the interplay between agent morphology and optimization constraints. Together, these contributions bridge a gap between theoretical insights and empirical evaluation, offering a foundation for more robust and morphology-aware policy optimization in high-dimensional control tasks.

While the study demonstrates a clear dependence of PPO’s clipping behavior on action space dimensionality and shows that compensation can restore fair treatment across dimensions from an information propagation perspective, it does not prove that this adjustment also yields uniformly performing policies across morphologies. Treating all dimensions equally in the gradient update does not necessarily account for the varying informational contribution each dimension makes from a meta-reinforcement learning viewpoint. A more complete solution would adapt update magnitudes not only to action space size but also to the relevance of each control dimension for task generalization.

Furthermore, the proposed compensation scheme is most effective in small to moderately sized action spaces, as commonly found in robotics where the number of controllable joints is limited. In such settings, scaling $\epsilon$ to account for action space dimensionality preserves stable learning while mitigating underestimation of the effective trust region, leading to faster convergence and improved sample efficiency. In contrast, for very large action spaces, the required scaling increases $\epsilon$ to values that undermine the trust region principle, reducing the method’s theoretical soundness and potential benefits. In those cases, standard PPO or alternative constraint formulations may remain preferable, as also evidenced by prior work.

A promising direction for future research is to investigate morphological generalization under alternative trust region constraints that are less sensitive to action space dimensionality. In particular, metrics such as the Wasserstein distance, discussed in Section~\ref{sec:related_work}, may offer dimension-invariant update constraints and could help align parameter updates with both structural and task-level information available to the agent.

\printbibliography

\end{document}